\documentclass[letterpaper]{article} 
\usepackage{aaai2026} 
\usepackage{times}  
\usepackage{helvet}  
\usepackage{courier}  
\usepackage[hyphens]{url}  
\usepackage{graphicx} 
\usepackage{natbib}  
\usepackage{caption} 
\usepackage{algorithm}
\usepackage{algorithmic}
\usepackage[table,xcdraw]{xcolor}
\usepackage{amsmath} 
\usepackage{xcolor}
\usepackage{graphicx}

\definecolor{brightpink}{RGB}{255,105,180}

\usepackage{newfloat}
\usepackage{listings}
\DeclareCaptionStyle{ruled}{labelfont=normalfont,labelsep=colon,strut=off} 
\floatstyle{ruled}
\newfloat{listing}{tb}{lst}{}
\floatname{listing}{Listing}
\title{HumanSense: From Multimodal Perception to Empathetic Context-Aware Responses through Reasoning MLLMs}
\author {
    Zheng Qin\textsuperscript{\rm 1,2},
    Ruobing Zheng\textsuperscript{\rm 2}$^{\dag}$,
    Yabing Wang\textsuperscript{\rm 1},
    Tianqi Li\textsuperscript{\rm 2},
    Yi Yuan\textsuperscript{\rm 2},
    Jingdong Chen\textsuperscript{\rm 2},
    Le Wang\textsuperscript{\rm 1}$^{*}$
}

\usepackage{bibentry}

\begin{document}

\maketitle

\footnotetext{$^\dag$Co-first author. Project lead. $^*$Corresponding author. $^1$National Key Laboratory of Human-Machine Hybrid Augmented Intelligence, National Engineering Research Center for Visual Information and Applications, Institute of Artificial Intelligence and Robotics, Xi'an Jiaotong University. $^2$Ant Group.}

\begin{abstract}
While Multimodal Large Language Models (MLLMs) show immense promise for achieving truly human-like interactions, progress is hindered by the lack of fine-grained evaluation frameworks for human-centered scenarios, encompassing both the understanding of complex human intentions and the provision of empathetic, context-aware responses. 
Here we introduce \textbf{HumanSense}, a comprehensive benchmark designed to evaluate the human-centered perception and interaction capabilities of MLLMs, with a particular focus on deep understanding of extended multimodal contexts and the formulation of rational feedback. 
Our evaluation reveals that leading MLLMs still have considerable room for improvement, particularly for advanced interaction-oriented tasks. Supplementing visual input with audio and text information yields substantial improvements, and Omni-modal models show advantages on these tasks.
Furthermore, grounded in the observation that appropriate feedback stems from a contextual analysis of the interlocutor's needs and emotions, we posit that reasoning ability serves as the key to unlocking it. We devise a multi-stage, modality-progressive reinforcement learning approach, resulting in \textbf{HumanSense-Omni-Reasoning}, which substantially enhances performance on higher-level understanding and interactive tasks. Additionally, we observe that successful reasoning processes appear to exhibit consistent thought patterns. By designing corresponding prompts, we also enhance the performance of non-reasoning models in a training-free manner.
Project page: \textcolor{brightpink}{\url{https://digital-avatar.github.io/ai/HumanSense/}}
\end{abstract}


\section{Introduction}
Science fiction~\cite{detroit2018,her2013} often portrays a future where artificial intelligence serves not merely as a tool for task execution, but as a human companion offering social support and emotional connection. The fundamental evolution from narrow, task-oriented systems to Artificial General Intelligence is predicated on the capacity to comprehend human intentions from speech, expressions, and body language, thereby enabling appropriate responses.

Multimodal Large Language Models (MLLMs)~\cite{Qwen2.5-Omni,hurst2024gpt,claude3_2024,gemini2023} represent a promising pathway toward realizing this vision. Their ability to holistically process visual, auditory, and textual information enables a comprehensive understanding of users and environments. MLLMs also have the potential to deeply analyze perceived information~\cite{guo2025deepseek} and subsequently plan appropriate feedback, which is not limited to textual responses, but can include suitable emotions, tones, and gesture labels in temporal sequences. Such outputs can be further integrated with video generation~\cite{meng2025echomimicv3,qin2025versatile}, speech synthesis~\cite{wang2023valle,bark2023}, and talking head~\cite{li2024ditto,li2024lokitalk,zhang2024learning,zheng2021neural} methods to provide a highly anthropomorphic interactive experience.

Achieving this goal first requires defining the necessary capabilities, evaluating model performance, and then applying optimization. However, existing benchmarks~\cite{qi2025vcr,lin2024streaming,hu2025video} lack targeted, fine-grained evaluation for these human-centered scenarios. To address this gap, we first define the primary capabilities needed for MLLMs in such scenarios: 1) multi-modal perception, 2) contextual understanding of implicit information, and 3) appropriate responses in multi-turn interactions. For interactive scenarios, we consider both response content and response strategies.

\begin{figure*}[t]
\centering
\includegraphics[width=1\textwidth]{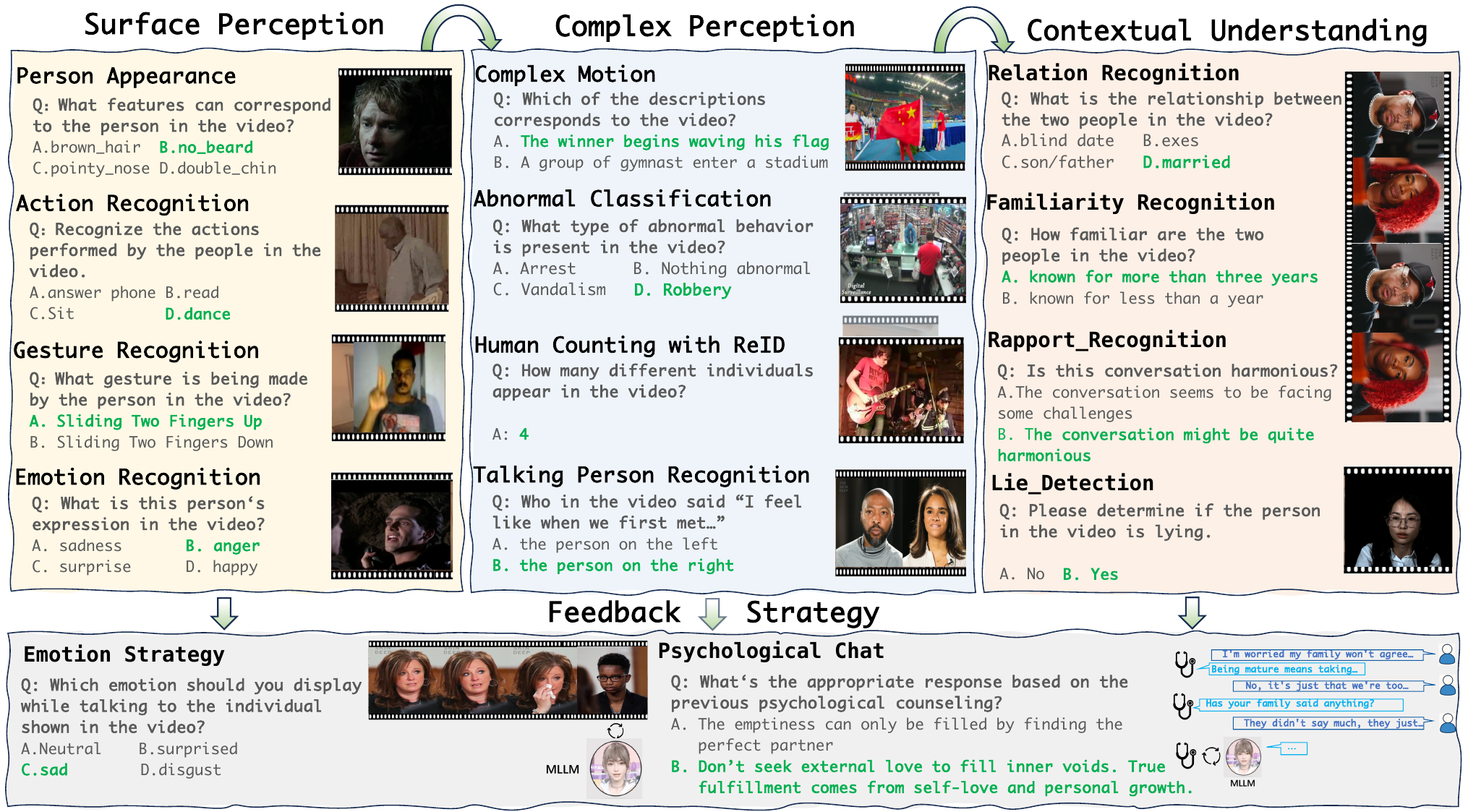}
\caption{HumanSense benchmark is structured hierarchically to evaluate the end-to-end, human-centric process of perception, understanding, and feedback through a series of demonstrated tasks and questions.}
\label{intro}
\end{figure*}

Based on the above considerations, we propose the HumanSense benchmark. This benchmark comprises 15 progressively challenging tests, totaling 3,882 questions derived from real-world records. In interactive tests, MLLMs are tasked with assuming the role of one party in the interaction and generating responses, which are then compared to real human records. We conduct a comprehensive evaluation of current leading MLLMs, including Vision-Language Models~\cite{zhu2025internvl3,qwen2.5-VL}, Omni models~\cite{hurst2024gpt,Qwen2.5-Omni}, and Audio-Language Models~\cite{Qwen-Audio}. The results reveal significant room for improvement in human-centered scenarios, particularly in advanced interaction-oriented tasks. Modality ablation studies show that visual, auditory, and textual information all play crucial roles in high-level tasks, and omni models capable of jointly processing audio, video, and text exhibit a clear advantage.

Building on these findings, we propose that omni-modal reasoning can enhance the cognitive and interactive capabilities of MLLMs. This assertion stems from our observation that appropriate feedback in communication relies on thorough consideration of omni-modal context, the interlocutor's needs, emotions, and personal characteristics. Accordingly, we employ a multi-stage, omni-modal reinforcement learning approach to build a reasoning omni model, resulting in substantial improvements in evaluations. Furthermore, we observe that successful reasoning processes exhibit highly consistent patterns. By designing corresponding prompts, we also enhance the performance of non-reasoning models in a training-free manner. 

The path to Artificial General Intelligence requires long-term, multifaceted exploration. With this work, we aim to inspire the community to explore the potential of Omni MLLMs for improving human-centered AI interactions and helps shape this emerging direction.

\section{Related Works}
\paragraph{Multimodal Large Language Models.}
LLMs~\cite{bai2023qwen,mann2020language,radford2018improving,radford2019language,touvron2023llama} have been extensively adopted for human behavior and sentiment analysis, facilitating applications including dialogue simulation, behavior prediction, and textual sentiment classification. These capabilities support diverse scenarios ranging from social media monitoring to automated customer support systems. However, textual information alone is often insufficient, as these models lack support for visual cues such as facial expressions and body language, which are essential for comprehensive human behavior analysis.
Visual MLLMs~\cite{lin2023video,chen2024sharegpt4video,li2024llava,wang2024qwen2,team2024gemini,yao2024minicpm} demonstrate strong capabilities in visual understanding, accurately recognizing emotions and behaviors through analysis of facial expressions and body language. However, a critical limitation of these models is their inability to process audio information, resulting in the loss of crucial auditory cues such as dialogue content, vocal intonations, and ambient sounds. This limitation creates significant gaps and introduces biases in their understanding of complex real-world scenarios.
Omni models~\cite{Qwen2.5-Omni,liu2025ola,zhang2024internlmxcomposer25omnilivecomprehensivemultimodallongterm,fu2025vita,fang2024llama,zhang2023speechgpt,hurst2024gpt}, by contrast, integrate multiple modalities—including vision, language, and audio—to provide comprehensive simulation of complex human interactions. These models can process dialogue content while simultaneously analyzing visual cues, enabling more nuanced and accurate understanding of human communication dynamics.

\paragraph{MLLM Benchmarks.}
With the advancement of multimodal large models, several evaluation benchmarks~\cite{chen2024sharegpt4video, fu2025video, wang2024lvbench, li2024omnibench, zhang2024flash} have emerged, most of which focus on assessing video understanding capabilities. Additionally, StreamingBench~\cite{lin2024streaming} specializes in streaming video understanding. However, few benchmarks evaluate large models from a human-centered perspective, which is crucial for the practical deployment of such models in real-world scenarios. 
HumanOmniV2~\cite{yang2025humanomniv2} focuses on deciphering intentions, interpreting emotions, and detecting potential deception in an omni-modal manner.
While HumanOmniV2 offers valuable insights into human-centered video understanding, it lacks the evaluation of the response planning or interactive capabilities. 

\begin{figure*}[t]
\centering
\includegraphics[width=1\textwidth]{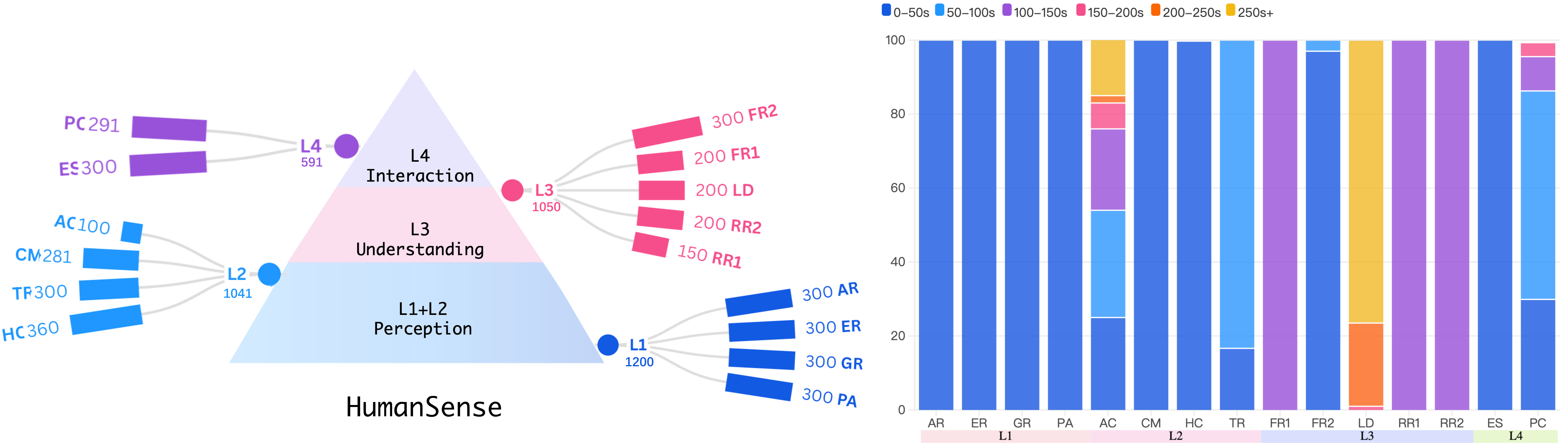} 
\caption{\textbf{Left}: The structure and counts of evaluation tasks in HumanSense. \textbf{Right}: Benchmark statistics on video length.}
\label{video_length}
\end{figure*}

\section{HumanSense-Bench}
\subsection{Overview}
We aim to systematically assess human-centered capabilities of MLLM through HumanSense framework: 1) Human-centered multi-modal perception, 2) Contextual understanding of implicit information, and 3) Response strategy in interactive scenarios, as Figure~\ref{intro}. The evaluation tasks are organized into a four-tier pyramid structure (L1–L4) according to increasing levels of difficulty, as shown in Figure~\ref{video_length} (Left):
\begin{itemize}
    \item \textbf{Perception} (L1 \& L2): The base layer focuses on unimodal, surface perception tasks in L1, and L2 addresses multi-modal, long-duration complex perception tasks. Together, they form the foundational capabilities of intelligence.
    \item \textbf{Understanding} (L3): Built upon perception, this layer evaluates whether the model can uncover implicit information embedded in conversations.
    \item \textbf{Response} (L4): As the pinnacle of capabilities, this layer assesses the model's ability to generate appropriate and rational responses in various interactive scenarios.
\end{itemize}
This design ensures the systematic nature of the evaluation: the model must first possess a solid perceptual foundation before advancing to deep understanding, ultimately enabling it to make informed feedback decisions at the top level. See Figure~\ref{video_length} for an overview of HumanSense tasks and dataset statistics on task quantity distribution and video length.

\subsection{Task Definition}
In human communication, diverse information is conveyed through different modalities. For example, visual expressions such as facial expressions and gestures can transmit emotional or semantic information; sound can directly express content or indirectly convey emotions. Perceiving these fundamental pieces of information is essential for interaction. We design the following multi-modal perception tasks.

\subsubsection{L1: Surface Perception}
\begin{itemize}

    \item \textbf{Person Appearance (PA)} evaluates the model's fine-grained perception of facial appearance, as appearance constitutes a fundamental aspect of person identification. We leverage the data annotations from \textit{CelebV-HQ}~\cite{zhu2022celebvhq} to design a series of multiple-choice questions. Each question asks whether the person in the video possesses a specific attribute, such as \textit{``Male"}, \textit{``Young"}, \textit{``Chubby"}, \textit{``Rosy cheeks"}, \textit{``Oval face"}, or \textit{``straight hair"}.

    \item \textbf{Action Recognition (AR)} aims to evaluate the model's ability to recognize atomic human actions. We ask MLLMs to identify the current action from a movie clip which comes from the \textit{AVA (Atomic Visual Actions)}~\cite{gu2018ava} dataset. The actions include individual behaviors (e.g., \textit{``walk"}, \textit{``sleep"}), interactions with objects (e.g., \textit{``Open a window"}, \textit{``Row a boat"}), and interactions between people (e.g., \textit{``Kiss a person"}, \textit{``talk to a person"}).

    \item \textbf{Gesture Recognition (GR)} aims to evaluate the model's ability to recognize hand gestures, which convey rich semantic information during communication. We construct single-choice questions using \textit{Jester}~\cite{materzynska2019jester} dataset, with gestures such as \textit{``Rolling Hand Forward"} and \textit{``Sliding Two Fingers Up"}.

    \item \textbf{Emotion Recognition (ER)} examines the recognition of facial expressions, as they are the primary means of conveying emotions. Based on \textit{CelebV-HQ}~\cite{zhu2022celebvhq} dataset, we generate single-choice questions using its built-in labels, asking the model to identify the emotion in videos such as \textit{``Happy"},\textit{``Sad"}, \textit{``Disgust"}, \textit{``Anger"}, \textit{etc.}

\end{itemize}

\subsubsection{L2: Complex Perception}

\begin{itemize}
  \item \textbf{Complex Motion (CM)} examines the description of extended complex action sequences, which relates to the understanding of target behaviors. We utilize the captions from \textit{ActivityNet} dataset~\cite{caba2015activitynet} to construct single-choice questions, where the model must identify the correct description of the action performed in the long video clip.

  \item \textbf{Abnormal Classification (AC)} evaluates the detection of abnormal human behaviors. We formulate single-choice questions based on the \textit{UCF-Crime100} dataset~\cite{sultani2018real}. The abnormal events include \textit{``Stealing"}, \textit{``Robbery"}, or \textit{``Fighting"}, \textit{etc.}

  \item \textbf{Human Counting with ReID (HC)} evaluates the model's ability to recognize and remember individuals. We use the tracking dataset \textit{TAO}~\cite{dave2020tao} to calculate the number of individuals and ask the model about the total count of distinct persons appearing throughout a video. Some challenging questions involve camera transitions and the intermittent appearance and disappearance of humans.
  
  \item \textbf{Talking Person Recognition (TR)} evaluates the ability to make judgments by integrating visually and auditorily perceived information. Based on \textit{RealTalk} dataset~\cite{geng2023affective}, we extract video clips and formulate single-choice questions, where the model is required to identify which person is speaking a specific content.

\end{itemize}

The L1 and L2 tasks assess the perceptual abilities of MLLMs in relation to ``seeing" and ``hearing". Achieving harmonious communication requires deep thinking about contextual content and providing appropriate responses that correspond to the interlocutor's emotions. Accordingly, L3 examines the model's capacity to ``understand" implicit information during interactions, while L4 evaluates the model's ``response" abilities across different scenarios.


\subsubsection{L3: Contextual Understanding}
\begin{itemize}
  \item \textbf{Familiarity Recognition (FR)} evaluates the model's ability to understand human interpersonal interactions and perceive interpersonal closeness. Based on the duration of acquaintance between conversational participants in \textit{The Skin Deep} channel, we construct single-choice questions to have the model determine the familiarity level between individuals in the videos.
  

  \item \textbf{Rapport Recognition (RR)} evaluates the model's ability to perceive whether the communication atmosphere is harmonious. We annotate video chat content from \textit{The Skin Deep} channel across multiple dimensions, including interaction frequency, communication atmosphere, and degree of viewpoint conflict, to construct single-choice questions. The evaluation encompasses whether the interactive atmosphere is pleasant and whether conflicts exist in the conversational content.
  

 \item \textbf{Relation Recognition (RG)} evaluates the model's ability to determine human relationships through multi-modal context. Based on the relationships between conversational participants in \textit{The Skin Deep} channel, we construct single-choice questions to have the model predict the relationship type between individuals in the videos, such as \textit{``Married"},\textit{``Siblings"},\textit{``Son/Mother"}, \textit{etc.}. The model must integrate information from visual appearance, age differences, and dialogue content to make a judgment.
 

\item \textbf{Lie Detection (LD)} evaluates the model's ability to detect human lying. We formulated single-choice questions based on the \textit{SEUMLD} dataset. The model is required to determine whether the speaker is lying in each video segment through analysis of visual and audio cues.


  \item \textbf{Fraud Recognition (FG)} evaluates the detection of specific fraudulent behaviors through approximately one-minute phone recordings, derived from the \textit{Telecom Fraud Texts} dataset~\cite{li2024innovative}. The fraud types include ``\textit{Loan and credit card agency fraud}", ``\textit{Impersonation of public security, judiciary, and government agencies}" and \textit{``Impersonation of leaders or acquaintances"}, \textit{etc.}
  

\end{itemize}

\subsubsection{L4: Feedback Strategy}
\begin{itemize}
  \item \textbf{Emo Strategy (ES)} evaluates the MLLM's ability to provide appropriate facial expression feedback during communication, creating an empathetic interaction experience. Based on the \textit{RealTalk} dataset~\cite{geng2023affective}, we extract video clips using the speaker as input for questions, and annotated the listener's facial expressions as answers to construct single-choice questions.
  

  \item \textbf{Psychological Chat (PC)} evaluates the ability of MLLMs to generate appropriate responses in complex, long-context interactions. Here we use a professional online psychological dataset, \textit{Emotional First Aid} dataset~\cite{efaqa-corpus-zh:petpsychology}, from which we constructed single-choice questions. The model is expected to select an appropriate response based on the previous multi-turn dialogue.


\end{itemize}


\section{Data Construction}
\paragraph{Question-Answer Generation.}
HumanSense consists of 3,291 video-based questions and 591 audio-based questions. The related data is sourced from existing open-source datasets and YouTube videos. We construct Question-Answer~(QA) pairs using templates, leveraging annotations and built-in labels from the existing datasets. We also use various off-the-shelf modules, including emotion recognition, Large Language Models(LLMs), and Optical Character Recognition~(OCR), to analyze source data and extract task-relevant information. Detailed information on the construction for each task is available in the supplementary materials.

\begin{table*}[t]
\centering
\resizebox{1\textwidth}{!}{
\begin{tabular}{l|ccccccccccccccccc}
Models                                                    & \textbf{Avg.}                 & \textbf{Avg.*} & AR                                           & ER                                           & GR                                           & PA                                           & AC                                           & CM                                           & HC                                           & TR                                           & FR                                           & FG                                           & LD                                           & RR                                           & RG                                           & ES                                           & PC                                           \\
                                                          &                               &                & \multicolumn{4}{c}{\cellcolor[HTML]{FBE4E7}L1}                                                                                                                                            & \multicolumn{4}{c}{\cellcolor[HTML]{FBDFEF}L2}                                                                                                                                            & \multicolumn{5}{c}{\cellcolor[HTML]{D9DFFC}L3}                                                                                                                                                                                           & \multicolumn{2}{c}{\cellcolor[HTML]{E8F7CF}L4}                                              \\ \hline
\multicolumn{1}{c|}{\textit{\textbf{}}}                   & \multicolumn{17}{c}{\textit{\textbf{HumanSense (tiny) Perf.}}}                                                                                                                                                                                                                                                                                                                                                                                                                                                                                                                                                                                                                                                                                                                  \\
Human Level $^\dagger$                                    & 0.875                         & 0.874          & 0.917                                        & 0.933                                        & 0.767                                        & 0.933                                        & 0.967                                        & 0.967                                        & 0.889                                        & 0.900                                        & 0.900                                        & 0.800                                        & 0.533                                        & 0.967                                        & 0.833                                        & 0.867                                        & 0.933                                        \\
GPT-4o$^\dagger$                                          & 0.552                         & -              & 0.583                                        & 0.233                                        & 0.700                                        & 0.517                                        & 0.733                                        & 0.767                                        & 0.522                                        & 0.400                                        & 0.833                                        & -                                            & 0.300                                        & 0.467                                        & 0.467                                        & 0.667                                        & -                                            \\
InternVL3-8B$^\dagger$                                    & 0.558                         & -              & 0.417                                        & 0.467                                        & 0.533                                        & 0.433                                        & 0.833                                        & 0.767                                        & 0.567                                        & 0.333                                        & 0.733                                        & -                                            & 0.667                                        & 0.433                                        & 0.433                                        & 0.633                                        & -                                            \\
Qwen2.5-Omni-7B$^\dagger$                                 & 0.578                         & 0.572          & 0.467                                        & 0.500                                        & 0.300                                        & 0.383                                        & 0.633                                        & 0.800                                        & 0.467                                        & 0.600                                        & 0.733                                        & 0.700                                        & 0.567                                        & 0.767                                        & 0.600                                        & 0.700                                        & 0.367                                        \\
Qwen2-Audio-7B$^\dagger$                                  & -                             & -              & -                                            & -                                            & -                                            & -                                            & -                                            & -                                            & -                                            & -                                            & -                                            & 0.333                                        & -                                            & -                                            & -                                            & -                                            & 0.333                                        \\ \hline
\multicolumn{1}{c|}{\textit{\textbf{}}}                   & \multicolumn{17}{c}{\textit{\textbf{HumanSense Perf.}}}                                                                                                                                                                                                                                                                                                                                                                                                                                                                                                                                                                                                                                                                                                                         \\
\rowcolor[HTML]{ECF4FF} \textit{Proprietary Models (API)} &                               &                &                                              &                                              &                                              &                                              &                                              &                                              &                                              &                                              &                                              &                                              &                                              &                                              &                                              &                                              &                                              \\
GPT-4o                                                    & 0.557                         & -              & 0.548                                        & 0.282                                        & \textbf{0.620}                               & 0.517                                        & \textbf{0.750}                               & \textbf{0.776}                               & 0.536                                        & 0.570                                        & 0.735                                        & -                                            & 0.310                                        & 0.480                                        & 0.535                                        & 0.587                                        & -                                            \\
\rowcolor[HTML]{ECF4FF} \textit{Vl-Model}                 &                               &                &                                              &                                              &                                              &                                              &                                              &                                              &                                              &                                              &                                              &                                              &                                              &                                              &                                              &                                              &                                              \\
LLaVA-Next-Video-7B                                       & 0.479                         & -              & 0.500                                        & 0.480                                        & 0.263                                        & 0.583                                        & 0.440                                        & 0.413                                        & 0.264                                        & 0.487                                        & 0.665                                        & -                                            & 0.505                                        & 0.560                                        & 0.500                                        & 0.577                                        & -                                            \\
Qwen2-VL-7B                                               & 0.507                         & -              & 0.473                                        & 0.470                                        & 0.307                                        & 0.322                                        & 0.600                                        & 0.591                                        & 0.424                                        & 0.537                                        & 0.665                                        & -                                            & 0.515                                        & 0.627                                        & 0.495                                        & 0.570                                        & -                                            \\
Qwen2.5-VL-7B                                             & 0.512                         & -              & 0.540                                        & 0.497                                        & 0.267                                        & 0.448                                        & 0.550                                        & 0.644                                        & 0.461                                        & 0.523                                        & 0.480                                        & -                                            & 0.545                                        & 0.627                                        & 0.485                                        & 0.590                                        & -                                            \\
VideoLLaMA3-7B                                            & 0.520                         & -              & 0.543                                        & 0.463                                        & 0.323                                        & 0.517                                        & 0.530                                        & 0.694                                        & 0.561                                        & 0.493                                        & 0.610                                        & -                                            & 0.515                                        & 0.587                                        & 0.400                                        & 0.530                                        & -                                            \\
LLaVA-OneVision-7B                                        & 0.521                         & -              & 0.545                                        & 0.510                                        & 0.400                                        & \textbf{0.592}                               & 0.620                                        & 0.676                                        & 0.268                                        & 0.503                                        & 0.600                                        & -                                            & 0.530                                        & 0.560                                        & 0.430                                        & 0.543                                        & -                                            \\
InternVL3-8B                                              & \textbf{0.561}                & -              & 0.393                                        & 0.483                                        & 0.387                                        & 0.547                                        & 0.670                                        & 0.751                                        & \textbf{0.630}                               & 0.567                                        & 0.735                                        & -                                            & 0.555                                        & 0.527                                        & 0.490                                        & 0.557                                        & -                                            \\
\rowcolor[HTML]{ECF4FF} \textit{Audio-Model}              &                               &                &                                              &                                              &                                              &                                              &                                              &                                              &                                              &                                              &                                              &                                              &                                              &                                              &                                              &                                              &                                              \\
Qwen2-Audio-7B                                            & -                             & -              & -                                            & -                                            & -                                            & -                                            & -                                            & -                                            & -                                            & -                                            & -                                            & 0.437                                        & -                                            & -                                            & -                                            & -                                            & 0.399                                        \\
\rowcolor[HTML]{ECF4FF} \textit{Omni-Model}               &                               &                & \multicolumn{1}{l}{\cellcolor[HTML]{ECF4FF}} & \multicolumn{1}{l}{\cellcolor[HTML]{ECF4FF}} & \multicolumn{1}{l}{\cellcolor[HTML]{ECF4FF}} & \multicolumn{1}{l}{\cellcolor[HTML]{ECF4FF}} & \multicolumn{1}{l}{\cellcolor[HTML]{ECF4FF}} & \multicolumn{1}{l}{\cellcolor[HTML]{ECF4FF}} & \multicolumn{1}{l}{\cellcolor[HTML]{ECF4FF}} & \multicolumn{1}{l}{\cellcolor[HTML]{ECF4FF}} & \multicolumn{1}{l}{\cellcolor[HTML]{ECF4FF}} & \multicolumn{1}{l}{\cellcolor[HTML]{ECF4FF}} & \multicolumn{1}{l}{\cellcolor[HTML]{ECF4FF}} & \multicolumn{1}{l}{\cellcolor[HTML]{ECF4FF}} & \multicolumn{1}{l}{\cellcolor[HTML]{ECF4FF}} & \multicolumn{1}{l}{\cellcolor[HTML]{ECF4FF}} & \multicolumn{1}{l}{\cellcolor[HTML]{ECF4FF}} \\
Ola-7B                                                    & \cellcolor[HTML]{FFFFFF}0.525 & 0.539          & \cellcolor[HTML]{FFFFFF}\textbf{0.557}       & \cellcolor[HTML]{FFFFFF}0.463                & \cellcolor[HTML]{FFFFFF}0.263                & \cellcolor[HTML]{FFFFFF}0.573                & \cellcolor[HTML]{FFFFFF}0.320                & \cellcolor[HTML]{FFFFFF}0.420                & \cellcolor[HTML]{FFFFFF}0.371                & \cellcolor[HTML]{FFFFFF}0.597                & \cellcolor[HTML]{FFFFFF}\textbf{0.785}       & \cellcolor[HTML]{FFFFFF}0.733                & \cellcolor[HTML]{FFFFFF}\textbf{0.565}       & \cellcolor[HTML]{FFFFFF}0.653                & \cellcolor[HTML]{FFFFFF}0.640                & \cellcolor[HTML]{FFFFFF}0.577                & \cellcolor[HTML]{FFFFFF}\textbf{0.567}       \\
IXC2.5-OmniLive-7B                                        & 0.544                         & 0.494          & 0.508                                        & 0.467                                        & 0.257                                        & 0.338                                        & 0.660                                        & 0.584                                        & 0.544                                        & 0.533                                        & 0.780                                        & 0.415                                        & 0.500                                        & 0.560                                        & 0.470                                        & 0.463                                        & 0.324                                        \\
Qwen2.5-Omni-7B                                           & 0.554                         & \textbf{0.559} & 0.473                                        & \textbf{0.513}                               & 0.303                                        & 0.350                                        & 0.600                                        & 0.630                                        & 0.438                                        & \textbf{0.600}                               & 0.770                                        & \textbf{0.740}                               & 0.550                                        & \textbf{0.713}                               & \textbf{0.650}                               & \textbf{0.607}                               & 0.399                                       
\end{tabular}
}
\caption{\textbf{Evaluation on HumanSense.} $^\dagger$ Indicates results on the HumanSense~(tiny) set, for comparison with human-level performance. While GPT-4o is designed as an omni-modal, the absence of audio input support in its current API precluded its evaluation on audio-related tasks~(FG, PC). 
We report two overall average scores: Avg., which excludes the two audio tasks to ensure a fair cross-model comparison, and Avg.*, which is the average across all tasks.}
\label{table1}
\end{table*}

\paragraph{Question-Answer Augmentation.}

To improve evaluation generalizability and avoid evaluation bias, we augment both questions and answers. For each questions, we design diverse candidate templates for random selection. For answers, we balance the distribution of correct options and equalize the lengths of correct and incorrect choices. This QA augmentation also avoid reward hacking~\cite{weng2024rewardhack} during the following Reinforcement Learning experiments.

\paragraph{Quality Control.}
We conduct manual inspection on a twenty percent sample of all QA pairs, with particular focus on L3 and L4 tasks. For example, in the ``Psychological Chat", we ensure that context contains sufficient multi-turn dialogue. We also rigorously validate the correct options to ensure they provide professional and appropriate advice. In the``Rapport Recognition” task, we design scoring dimensions including interaction frequency, communication atmosphere, and content harmony for LLMs evaluation, followed by manual quality inspection of the results.



\begin{figure}[t]
\centering
\includegraphics[width=0.9\columnwidth]{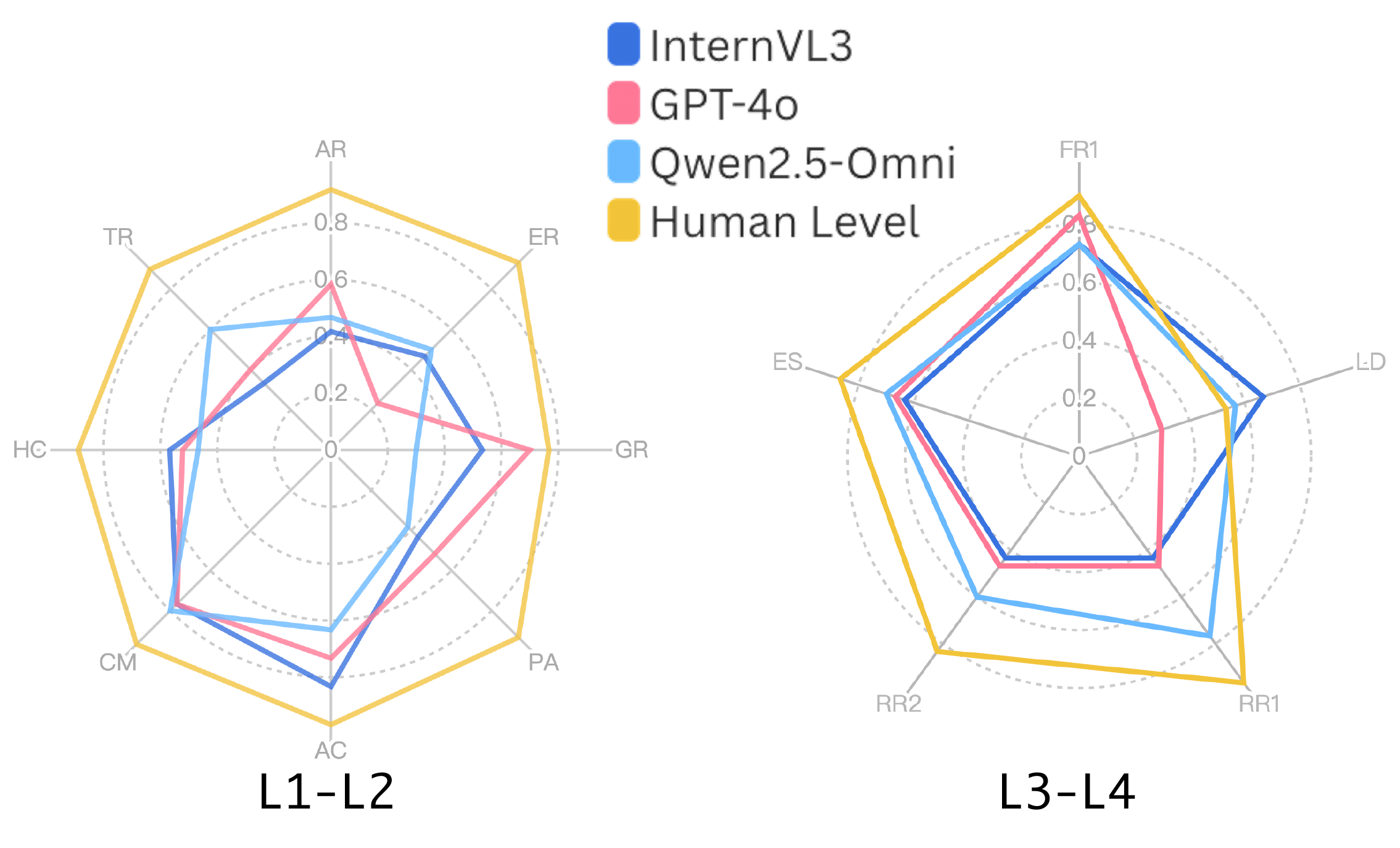}
\caption{\textbf{Performance Radar Charts on HumanSense (mini)}. The results include human-level performance with several state-of-the-art multimodal models.}
\label{sotas}
\end{figure}

\section{Evaluation on HumanSense}
We conduct a comprehensive evaluation of leading Multimodal Large Language Models (MLLMs) with sizes up to 10B, including: (1) Visual LLMs, which represent the most mainstream branch of MLLMs today; (2) Audio LLMs; and (3) Omni-modal LLMs that are natively designed for integrating vision, audio, and text. For Visual LLMs, we assess models such as Qwen2.5-VL~\cite{qwen2.5-VL}, Qwen2-VL~\cite{Qwen2VL}, LLaVA-OneVision~\cite{li2024llava}, LLaVANeXT-Video~\cite{liu2024llavanext}, VideoLLaMA3~\cite{zhang2025videollama} and InternVL3~\cite{zhu2025internvl3}. In the omni-modal models, we test Qwen2.5-Omni~\cite{Qwen2.5-Omni}, IXC2.5OmniLive~\cite{zhang2024internlmxcomposer25omnilivecomprehensivemultimodallongterm} and Ola~\cite{liu2025ola}. For audio LLMs, we evaluate Qwen2-Audio~\cite{Qwen-Audio}. Additionally, we test the powerful omni-model GPT-4o~\cite{hurst2024gpt}.
All evaluations are conducted in a zero-shot setting, employing the default prompts provided for each model. For video processing, we adhere to the respective official configurations for key parameters, including frame extraction methods, frames per second (FPS), and resolution. We release our data and code to the community, aiming to facilitate broader evaluation across a diverse range of models.

To accommodate the characteristics of each model, Visual LLMs are exclusively evaluated on video tasks. Similarly, Audio LLMs are assessed solely on audio tasks. In contrast, omni-models, which are designed for multimodal processing, are required to complete all tasks.

\paragraph{Human Level Performance.}
To establish a human performance benchmark, we curate a new evaluation set, dubbed HumanSense (tiny), comprising 450 questions randomly sampled across our tasks (30 per task). We first establish a human baseline by having evaluators independently answer each question. Subsequently, we compare the performance of several leading models—including GPT-4o~\cite{hurst2024gpt}, Intern3-VL~\cite{zhu2025internvl3}, Qwen2-Audio~\cite{Qwen-Audio}, and Qwen2.5-Omni~\cite{Qwen2.5-Omni} with human standard.


\begin{table*}[t]
\centering
\resizebox{0.65\textwidth}{!}{
\begin{tabular}{l|ccccccccc}
          & $ \mathrm{Avg}_{L1} $                 & $ \mathrm{Avg}_{L2} $                  & FR1            & FR2            & LD             & RR1            & RR2            & ES                     & PC                    \\
Models    & \cellcolor[HTML]{FBE4E7}L1 & \cellcolor[HTML]{FBDFEF}L2 & \multicolumn{5}{c}{\cellcolor[HTML]{D9DFFC}L3}                                     & \multicolumn{2}{c}{\cellcolor[HTML]{E8F7CF}L4} \\ \hline
Baseline  & 0.410                      & 0.567                      & 0.770          & 0.740          & 0.550          & 0.713          & 0.650          & 0.607                  & 0.399                 \\
+ Stg.1   & 0.555                      & 0.548                      & 0.720          & 0.557          & 0.540          & 0.707          & 0.620          & 0.593                  & 0.540                 \\
+ Stg.1-2 & 0.554                      & 0.565                      & 0.775          & 0.687          & 0.545          & 0.693          & 0.625          & 0.593                  & \textbf{0.625}        \\
+ Stg.1-3 & \textbf{0.563}             & \textbf{0.603}             & 0.780          & 0.687          & \textbf{0.555} & \textbf{0.720} & \textbf{0.690} & \textbf{0.620}         & 0.619                 \\ \hline
+ PE      & -                          & -                          & \textbf{0.790} & \textbf{0.763} & 0.523          & \textbf{0.720} & 0.625          & 0.600                  & 0.436                
\end{tabular}
}
\caption{\textbf{Evaluation of Multi-Stage Omni-Modal Reinforcement Learning and Training-Free Prompt Enhancement.} We report the average scores for L1 and L2, as well as detailed scores for each high-level task. PE represents Prompt Enhancement.}
\label{table2}
\end{table*}

\begin{figure}[t]
\centering
\includegraphics[width=1\columnwidth]{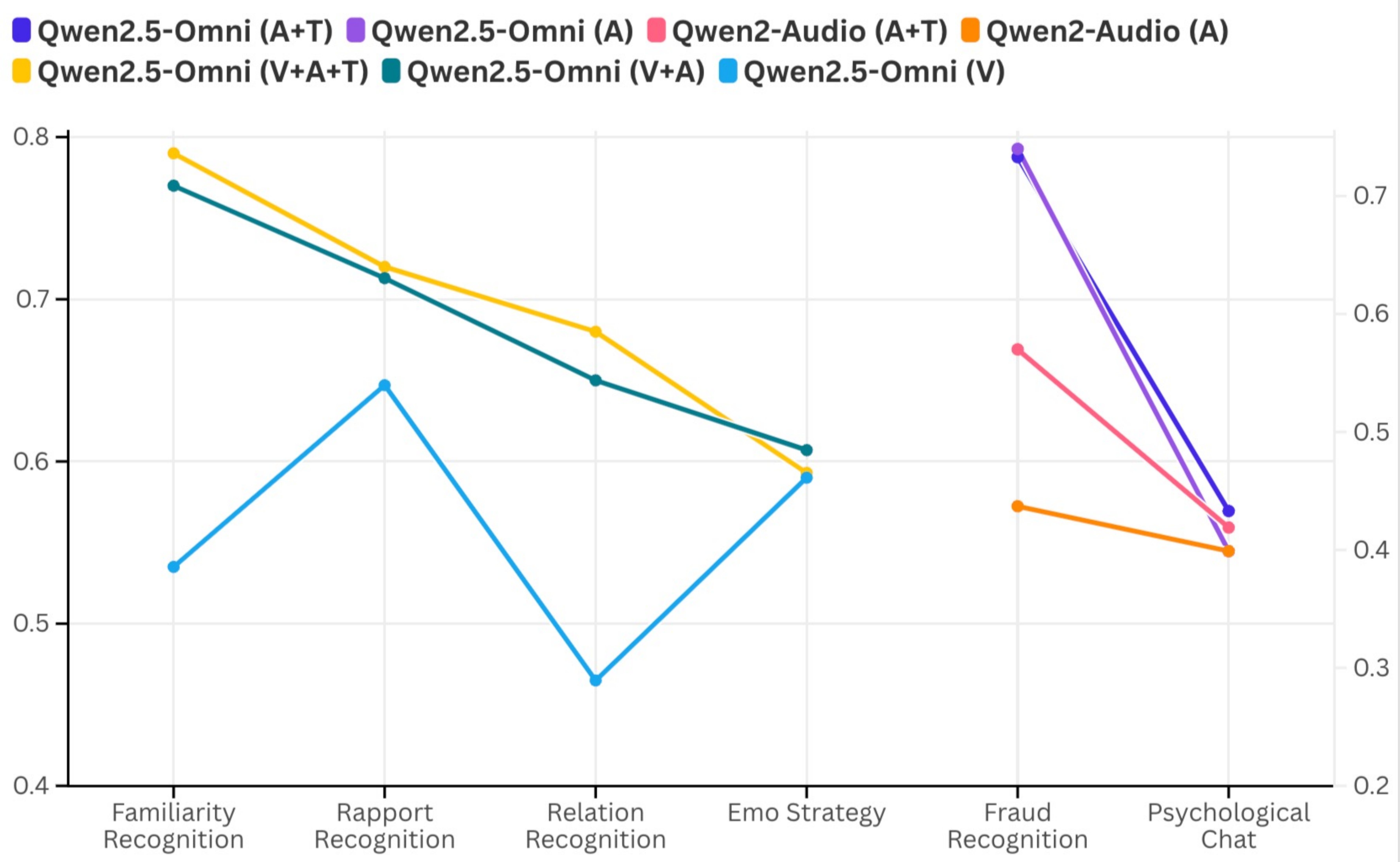}
\caption{\textbf{Modality Ablation Across Different Tasks.}We visualize the contributions of different modalities across tasks ranging from perception to interaction. The left 4 tasks are video-based questions, and the right 2 tasks are audio-based questions. Note that ASR-transcribed text (T) was used exclusively for this ablation study, not in Table~\ref{table1}.}
\label{model_ablation}
\end{figure}

\subsection{Results}

\paragraph{Human Level Performance.} Human evaluators achieve an average accuracy of 87.5\% on our benchmark, outperforming the best-performing model by a margin of 29.7\%. As shown in Figure~\ref{sotas}, a substantial performance gap still exists between all models and human-level performance, especially in complex L3-L4 tasks, highlighting a significant need for improvement in the capabilities of current MLLMs on human-centered tasks.

\begin{figure*}[t]
\centering
\includegraphics[width=1\textwidth]{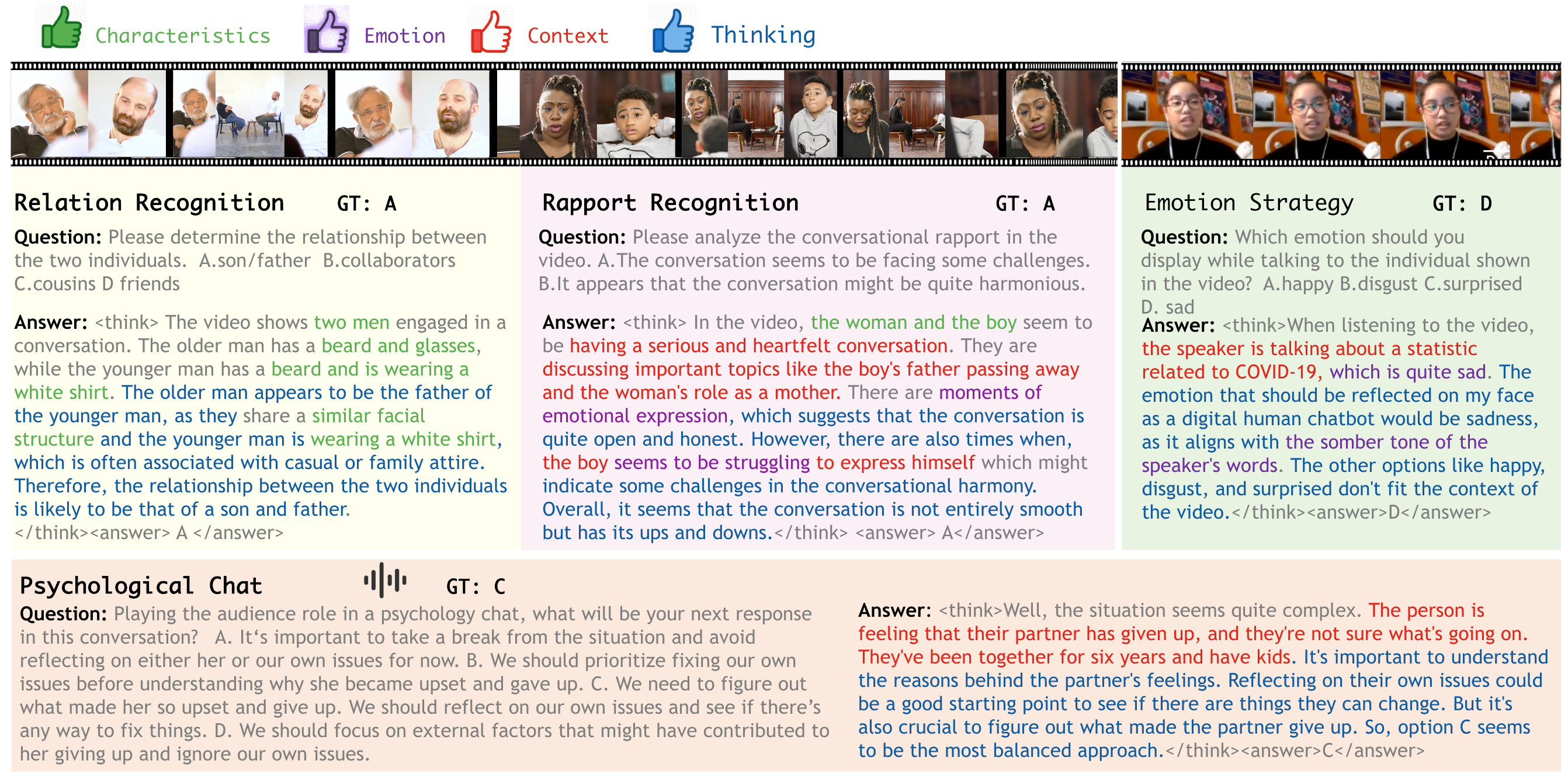}
\caption{\textbf{Examples of Successful Reasoning.} These cases cover four high-level perception and interaction tasks, including both video-based and audio-based questions. The reasoning processes all demonstrate thinking that integrates characteristics, emotions, and context, and then provides appropriate feedback.}
\label{visual1}
\end{figure*}

\paragraph{Visual LLMs.}
As Table~\ref{table1}, Intern3-VL shows the most advantages in terms of average performance, excelling in both L1-L2 perceptual tasks and L3-L4 high-level tasks. Specifically, LLaVA-OneVision achieves the highest metrics in L1 series tasks, reflecting its exceptional visual perception ability on basic tasks. InternVL3 stands out in L2 and some L3 tasks, demonstrating strong long-video memory and contextual understanding, which is consistent with its outstanding performance in video-related benchmarks~\cite{zhu2025internvl3}. Notably, in high-level L3-L4 tasks, all models exhibit metrics ranging from 40 to 60, with little variation, suggesting that visual modality input alone is insufficient to provide adequate information for these tasks.

\paragraph{Omni-models and Audio LLMs.}
The inclusion of audio grants omni-models a significant edge over visual-only LLMs in high-level tasks (L3, L4), such as Rapport Recognition and Lie Detection. This cross-modal synergy is further underscored in the Fraud Recognition task, where the Qwen2.5-Omni (0.74) outperforms its specialized audio counterpart. However, this perceptual advantage diminishes in more complex tasks like Psychological Chat. This highlights a limitation for current leading omni-models: the primary bottleneck is not low-level perception but rather a deficiency in high-level, long-context reasoning, which is essential for truly human-centered understanding.

The charts in Figure~\ref{sotas} visually demonstrate the omni-models, particularly Qwen2.5-Omni, show a marked advantage over the visual-only LLM in higher-level tasks, highlighting the critical role of multimodal, including auditory, perception.

\subsection{Modality Ablation}
To conduct a fine-grained analysis of the importance of different modalities, we performed a series of ablation studies using the Qwen2.5-Omni and Qwen2-Audio models. For clarity, we denote video, audio, and the ASR-transcribed text from video as V, A, and T, respectively. Our experiments focused on six challenging tasks from our L3-L4 difficulty tiers, comprising four video-based and two audio-based tasks. We designed two specific experimental settings to probe modal contributions: (1) augmenting both models with ASR-transcribed text (T) as an additional input, and (2) evaluating the Omni-model in a visual-only setting, removing the audio input.

Figure~\ref{model_ablation} demonstrates the results for our modality ablation study. 
For video tasks, audio input (A) serves as a powerful supplement to video (V) for the omni-model, significantly enhancing performance on high-level tasks such as rapport and relationship recognition. Incorporating ASR text (T) provides minimal additional benefit, indicating that raw audio already delivers substantial semantic information for this advanced model. In contrast, for audio-based tasks, the audio-only model exhibits a clear dependence on textual input, highlighting its limited speech comprehension and reliance on explicit semantic support. However, the omni-model gains little from text inputs, demonstrating the advantage of comprehensive multi-modal training. Furthermore, all models perform worse on response-related tasks compared to perceptual tasks, highlighting the importance of improving the response capabilities of MLLMs in interactive scenarios.


\section{From Perception to Responses}
The above evaluations confirm that visual, auditory, and textual information all play important roles in high-level tasks. Through data observation, we also find that appropriate feedback in communication relies on thorough consideration of omni-modal context and insight into the interlocutor's needs, emotions, and personal characteristics. Therefore, we believe that reasoning capabilities based on omni-modal inputs are key to enhancing the cognitive and interactive capabilities of MLLMs. In the following sections, we employ a multi-stage, omni-modal reinforcement learning approach to build a reasoning omni model.

\paragraph{Omni-modal Reinforcement Learning}

We construct a training set using data independent from the benchmark. We apply Group Relative Policy Optimization (GRPO)~\cite{shao2024deepseekmath} and design a multi-stage, omni-modal training approach that exposes the Qwen2.5-Omni-7B to all modalities during the Reinforcement Learning with Verifiable Reward (RLVR), enhancing training stability and progressively strengthening perceptual capabilities.

Specifically, in the first stage, we train using pure video frames and textual Question-Answer(QA) pairs to establish reasoning capabilities that integrate visual perception. In the second stage, we train with audio-based QA to develop reasoning that incorporates auditory perception. Finally, we utilize complete video-audio QA to reinforce reasoning that combines both visual and auditory perception. \textbf{Further details, such as reward functions, data sizes, and key hyperparameters, are provided in the Appendix. The complete training configurations are available in our released code.}

As shown in Table~\ref{table2}, for vision-centric L1 tasks, stage-1 training already yields significant improvements, indicating that reasoning grounded in visual perception can enhance performance on such tasks. For the audio-related tasks PC and FG, stage-2 training leads to notable gains over stage-1, revealing the success of incorporating auditory reasoning. Most tasks achieve optimal performance after completing all 3 training stages. We sample correctly answered examples and find that the model is indeed capable of deep thinking by integrating characteristics, emotions, and contextual information, as shown in Figure~\ref{visual1}. 

\paragraph{Training-Free Prompt Enhancement}
We observe that the successful reasoning processes elicited by RL training exhibit a consistent thought pattern: perceiving key characteristics, emotion, and context, followed by thinking and then response. Inspired by this, we believe there exists a training-free approach that can improve existing MLLMs' performance through prompt enhancement. To this end, we design and test the following prompt template, and find that it also boosts performance on high-level tasks, as shown in Table~\ref{table2}.
\begin{figure}[H]
\centering
\includegraphics[width=1\columnwidth]{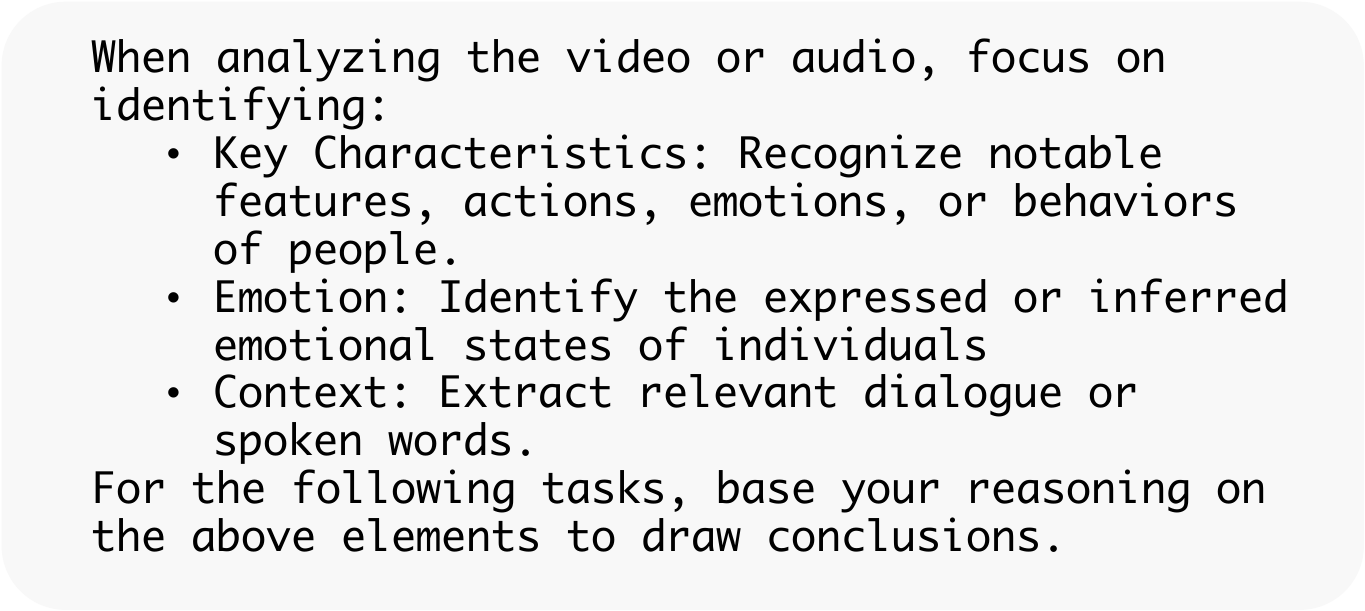}
\end{figure}

\section{Conclusion}
We introduce the HumanSense benchmark to explore MLLMs' capabilities in complex human-centered perception and interaction scenarios. We propose that omni-modal reasoning can enhance MLLMs' performance on such tasks. We aim to inspire the community to recognize the potential of MLLMs in advancing AI interaction experiences.

\section{Appendix}
\subsection{TRAINING DETAILS}
\label{sec:Training Details}
All three training stages are conducted on 8 H20 GPUs. Two types of rewards are employed: \textbf{Format Reward} and \textbf{Accuracy Reward}.

\begin{itemize}
    \item \textbf{Format Reward} incentivizes the model to generate predefined structural and labels. It gives a reward of 0 or 1 based on whether the output contains the complete tags of \textless think\textgreater\textless/think\textgreater.
 and \textless answer\textgreater\textless/answer\textgreater.

    \item \textbf{Accuracy Reward} evaluates the correctness of the outputs. It provides direct feedback based on how well the model's predictions align with ground truth or expected results.
\end{itemize}

Both rewards are assigned equal weight. The learning rate is set to $1 \times 10^{-6}$, and the gradient accumulation step is set to 1.

The details of the three training stages are as follows:
\begin{itemize}
    \item \textbf{Stage 1}: 2,663 video-only samples (without audio) are used for training over 400 steps.
    \item \textbf{Stage 2}: 1,000 audio-only samples are used for training over 150 steps.
    \item \textbf{Stage 3}: 1,283 multimodal samples (video and audio) are used for training over 200 steps.
\end{itemize}

Detailed scores for each task are presented in Table~\ref{table3}.

\subsection{EVALUATION PROTOCALS}
During large model evaluation, besides video or audio inputs, we employ the following prompt to guide the model in answering multiple-choice questions:
\begin{figure}[H]
\centering
\includegraphics[width=1\columnwidth]{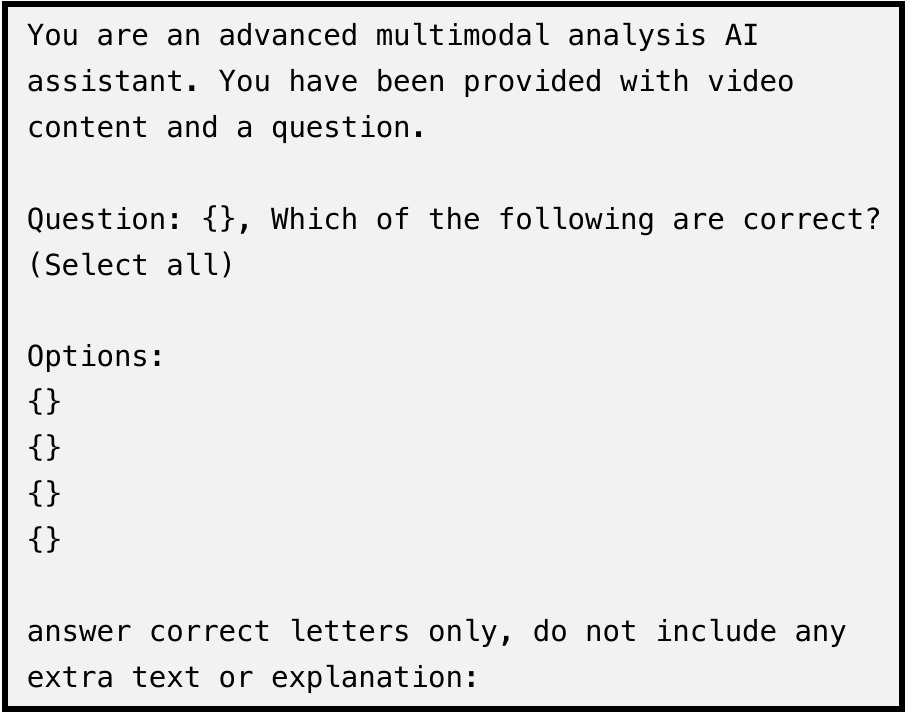}
\end{figure}
And we use the following prompt to have the model perform one-choice question:
\begin{figure}[H]
\centering
\includegraphics[width=1\columnwidth]{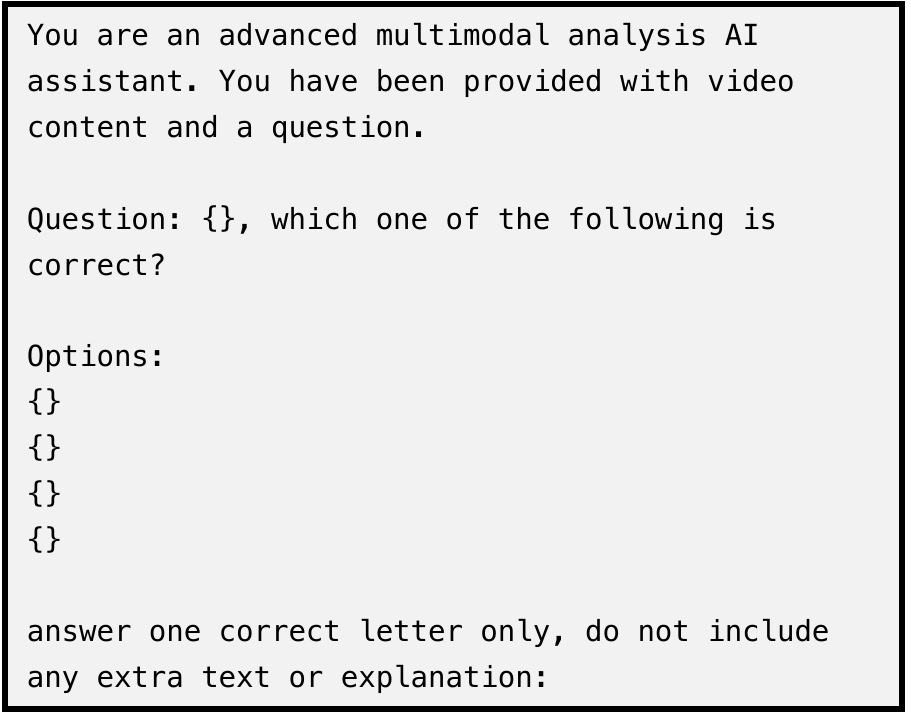}
\end{figure}
And we use the following prompt to have the model perform open-ended question~(Human counting with Reid):
\begin{figure}[H]
\centering
\includegraphics[width=1\columnwidth]{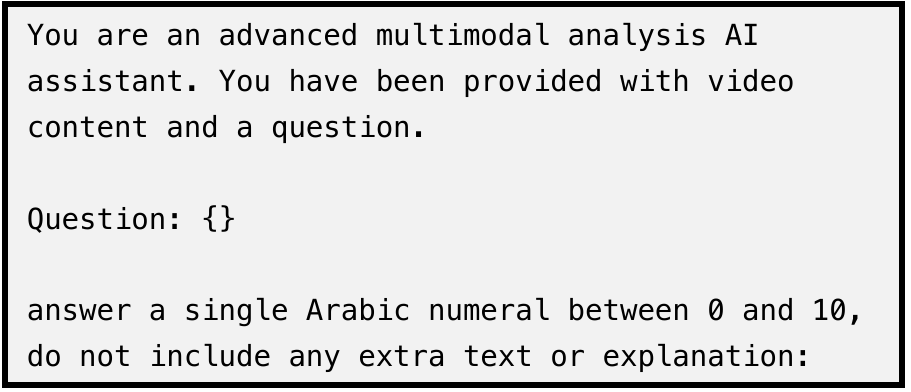}
\end{figure}

For the trained model, we augment each question with a prompt designed to stimulate its reasoning process, which is as follows:
\begin{figure}[H]
\centering
\includegraphics[width=1\columnwidth]{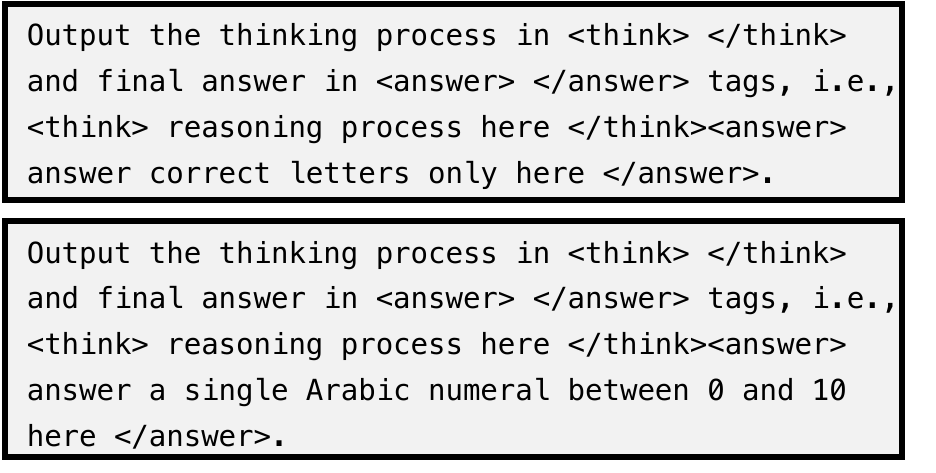}
\end{figure}

\subsection{MORE DETAILS OF DATA CONSTRUCTION}
For tasks such as Person Appearance, Action Recognition, Gesture Recognition, Fraud Recognition, Emotion Recognition, Abnormal Classification, and Lie Detection, we construct multiple-choice questions based on the built-in annotations from the original dataset (CelebV-HQ, AVA, Jester, Telecom Fraud Texts, CelebV-HQ, ActivityNet, UCF-Crime100, and SEUMLD).

For a single question, the video clip's corresponding ground truth (GT) label serves as the correct option, while distractor options are randomly sampled from the remaining choices in the answer set. A multiple-choice question is then constructed by combining the video with the correct option and the sampled distractors.


\paragraph{Human Counting with ReID.}
We generate question-answer (QA) pairs using the multi-object tracking dataset TAO. We filter the TAO videos to exclude those without human subjects, then segment the remaining videos into 30-50 second clips. 

For each segment, we identify the corresponding annotations from the original labels and calculate the total number of unique individuals present. This count serves as the numerical answer for our designed task, forming a complete question-answer pair.

\paragraph{Talking Person Recognition.}
We construct questions based on videos from RealTalk. We filter the clips by discarding segments where only one person speaks continuously. For each selected video that contains a dialogue, we extract a single quote. This quote is then used to pose the question: ``Who said this?"
The objective of this task is to test a large model's multimodal comprehension ability, specifically its capacity to synchronize audio with visual information to identify the speaker.

\paragraph{Relation Recognition.}
The show "The Skin Deep" displays title cards at the beginning of each episode, indicating the relationship between the two speakers. To create our question-answer pairs, we process the introductory segment of each video frame-by-frame using Optical Character Recognition (OCR) to extract this relationship information, serving as the ground truth for the question: ``What is the relationship between the two people in the video?"

\paragraph{Familiarity Recognition.}
At the beginning of ``The Skin Deep" episodes, on-screen text indicates the duration of the two speakers' acquaintance. This text serves as the ground truth for the question: "What is the level of familiarity between the two individuals in the video?"

\paragraph{Rapport Recognition.}
We segment clips from ``The Skin Deep" into 100-150 second lengths, then conduct a two-level assessment of each segment using both the video and its corresponding transcripts. The first level of video assessment evaluates the conversational interaction based on the following criteria:

\begin{itemize}
    \item \textbf{Interactivity:} The dialogue should feature frequent back-and-forth between participants. Points are deducted for long, awkward silences or extended pauses for thought.
   \item \textbf{Emotion:} Positive and shared emotions, such as happiness, humor, and relaxation—along with clear emotional resonance between the speakers are scored positively. Divergent or mismatched emotional states between participants result in a lower score.
    \item \textbf{Balance:}    Contributions to the conversation should be roughly equal from both parties. Scenarios where one individual dominates the conversation while the other is primarily a passive listener are penalized.
    \item \textbf{Coherence:} The discussion should remain focused on a consistent topic. Frequent and abrupt switching between topics will negatively impact the score. 

\end{itemize}
Furthermore, we employ large language models to directly rate the harmony of the ASR transcripts, identifying any disagreements.

\paragraph{Emo Strategy.}
We use pre-segmented two-person conversation clips from the RealTalk dataset, which are originally used for training speaker generation models. For our task, we select the ``talker" from each clip as the input. The corresponding ground truth label is the ``listener"s' facial expression, which we determine using a facial expression recognition algorithm. We refer to this as our 'emotion strategy' target.

\paragraph{Psychological Chat.}
We first filter high-quality advice and responses from online dialogue data of psychologists. This is done using the following prompt:
\begin{figure}[H]
\centering
\includegraphics[width=1\columnwidth]{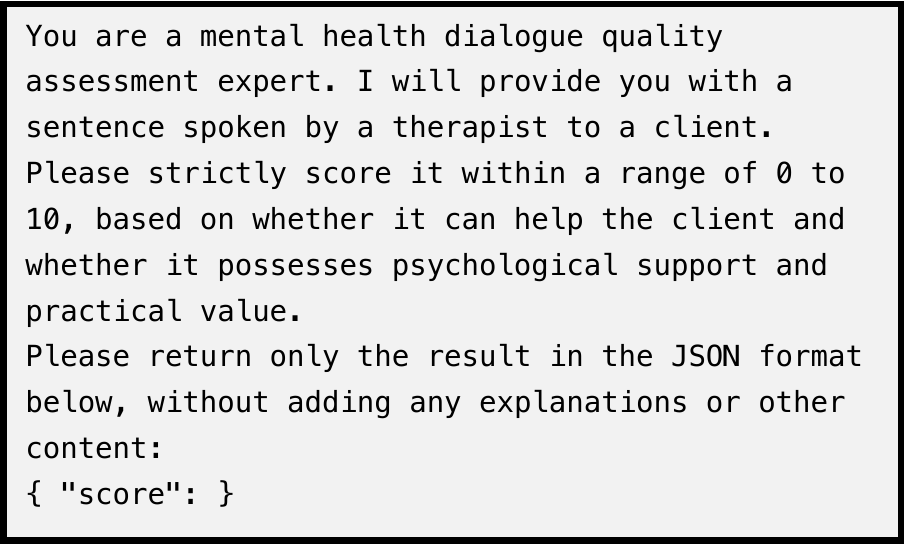}
\end{figure}
For the selected responses, we trace back more than 10 dialogue turns to include in the question, aiming to provide sufficient contextual information. Meanwhile, we reference the high-quality responses and generate distractor options.

\begin{table*}[t]
\centering
\resizebox{1\textwidth}{!}{
\begin{tabular}{lllllllllllllllll}
\hline 
Models            & \textbf{Avg.} & AR    & ER    & GR    & PA    & AC    & CM    & HC    & TR    & FR1   & FR2   & LD    & RR1   & RR2   & ES    & PC    \\\hline 
Baseline          & 0.559         & 0.473 & 0.513 & 0.303 & 0.350 & 0.600 & 0.630 & 0.438 & 0.600 & 0.770 & 0.740 & 0.550 & 0.713 & 0.650 & 0.607 & 0.399 \\
First Stage (v)   & 0.584         & 0.693 & 0.493 & 0.427 & 0.608 & 0.530 & 0.626 & 0.438 & 0.597 & 0.720 & 0.557 & 0.540 & 0.707 & 0.620 & 0.593 & 0.540 \\
Second Stage (a)  & 0.601         & 0.700 & 0.510 & 0.407 & 0.600 & 0.550 & 0.630 & 0.492 & 0.587 & 0.775 & 0.687 & 0.545 & 0.693 & 0.625 & 0.593 & 0.625 \\
Third Stage (v+a) & 0.619         & 0.705 & 0.513 & 0.420 & 0.615 & 0.550 & 0.633 & 0.488 & 0.740 & 0.780 & 0.687 & 0.555 & 0.673 & 0.690 & 0.620 & 0.619\\\hline 
\end{tabular}
}
\caption{Evaluation of Optimization Results based on Qwen2.5-Omni-7B. We provide the scores for each task.}
\label{table3}
\end{table*}



\subsection{Supplement the section on MLLM benchmarks in related works}

With the development of multimodal large models, their influence has permeated various fields~\cite{qin2025embracing, wang2025refdetector, wang2024referencing, wang2025mapping, li2025visual,wang2022cross}, leading to the emergence of numerous evaluation benchmarks~\cite{chen2024sharegpt4video,fu2025video,wang2024lvbench,li2024omnibench,zhang2024flash} have emerged. While they have been successful in general visual understanding, they rarely focus on the subtleties of human perception, comprehension, and planning. In the field of human behavior analysis, there are some specialized datasets. For example, in emotion recognition, AffectNet~\cite{mollahosseini2017affectnet} and in action recognition, Kinetics~\cite{smaira2020short}. While these datasets are crucial, they are typically designed for specific classification or detection tasks and do not provide a framework for evaluating a model's comprehensive reasoning ability. HumanOmniV2~\cite{yang2025humanomniv2} focuses on deciphering intentions, interpreting emotions, and detecting potential deception from video.
In summary, existing benchmarks either lack multimodal information, focus on general scenarios rather than human-centric ones, or are too task-specific. To systematically evaluate the capabilities of large models in complex human-centered scenarios, we have developed HumanSense.

\bigskip
\bibliography{aaai2026}

\setlength{\leftmargini}{20pt}
\makeatletter\def\@listi{\leftmargin\leftmargini \topsep .5em \parsep .5em \itemsep .5em}
\def\@listii{\leftmargin\leftmarginii \labelwidth\leftmarginii \advance\labelwidth-\labelsep \topsep .4em \parsep .4em \itemsep .4em}
\def\@listiii{\leftmargin\leftmarginiii \labelwidth\leftmarginiii \advance\labelwidth-\labelsep \topsep .4em \parsep .4em \itemsep .4em}\makeatother

\setcounter{secnumdepth}{0}
\renewcommand\thesubsection{\arabic{subsection}}
\renewcommand\labelenumi{\thesubsection.\arabic{enumi}}

\newcounter{checksubsection}
\newcounter{checkitem}[checksubsection]

\newcommand{\checksubsection}[1]{%
  \refstepcounter{checksubsection}%
  \paragraph{\arabic{checksubsection}. #1}%
  \setcounter{checkitem}{0}%
}

\newcommand{\checkitem}{%
  \refstepcounter{checkitem}%
  \item[\arabic{checksubsection}.\arabic{checkitem}.]%
}
\newcommand{\question}[2]{\normalcolor\checkitem #1 #2 \color{blue}}
\newcommand{\ifyespoints}[1]{\makebox[0pt][l]{\hspace{-15pt}\normalcolor #1}}

\end{document}


\maketitle

\section{Training Details}
\label{sec:Training Details}
All three training stages are conducted on 8 H20 GPUs, with a batchsize of 8. Two types of rewards are employed: \textbf{Format Reward} and \textbf{Accuracy Reward}.

\begin{itemize}
    \item \textbf{Format Reward}: This reward component incentivizes the model to generate outputs that conform to predefined structural and formatting standards. It evaluates the adherence of the model's output to expected formatting conventions, ensuring consistency and standardization in the generated content's presentation and organization.
    
    \item \textbf{Accuracy Reward}: This reward component evaluates the model's predictive performance by measuring the correctness of its outputs. It provides direct feedback based on how well the model's predictions align with ground truth or expected results.
\end{itemize}

Both rewards are assigned equal weight, ensuring that the model is equally incentivized to produce well-formatted outputs and to perform tasks with high accuracy. The learning rate is set to $1 \times 10^{-6}$, and the gradient accumulation step is set to 1.

The details of the three training stages are as follows:
\begin{itemize}
    \item \textbf{Stage 1}: 2,663 video-only samples (without audio) are used for training over 400 steps.
    \item \textbf{Stage 2}: 1,000 audio-only samples are used for training over 150 steps.
    \item \textbf{Stage 3}: 1,283 multimodal samples (video and audio) are used for training over 200 steps.
\end{itemize}

Detailed scores for each task are presented in Table~\ref{table3}.

\begin{figure}[t]
\centering
\includegraphics[width=1\columnwidth]{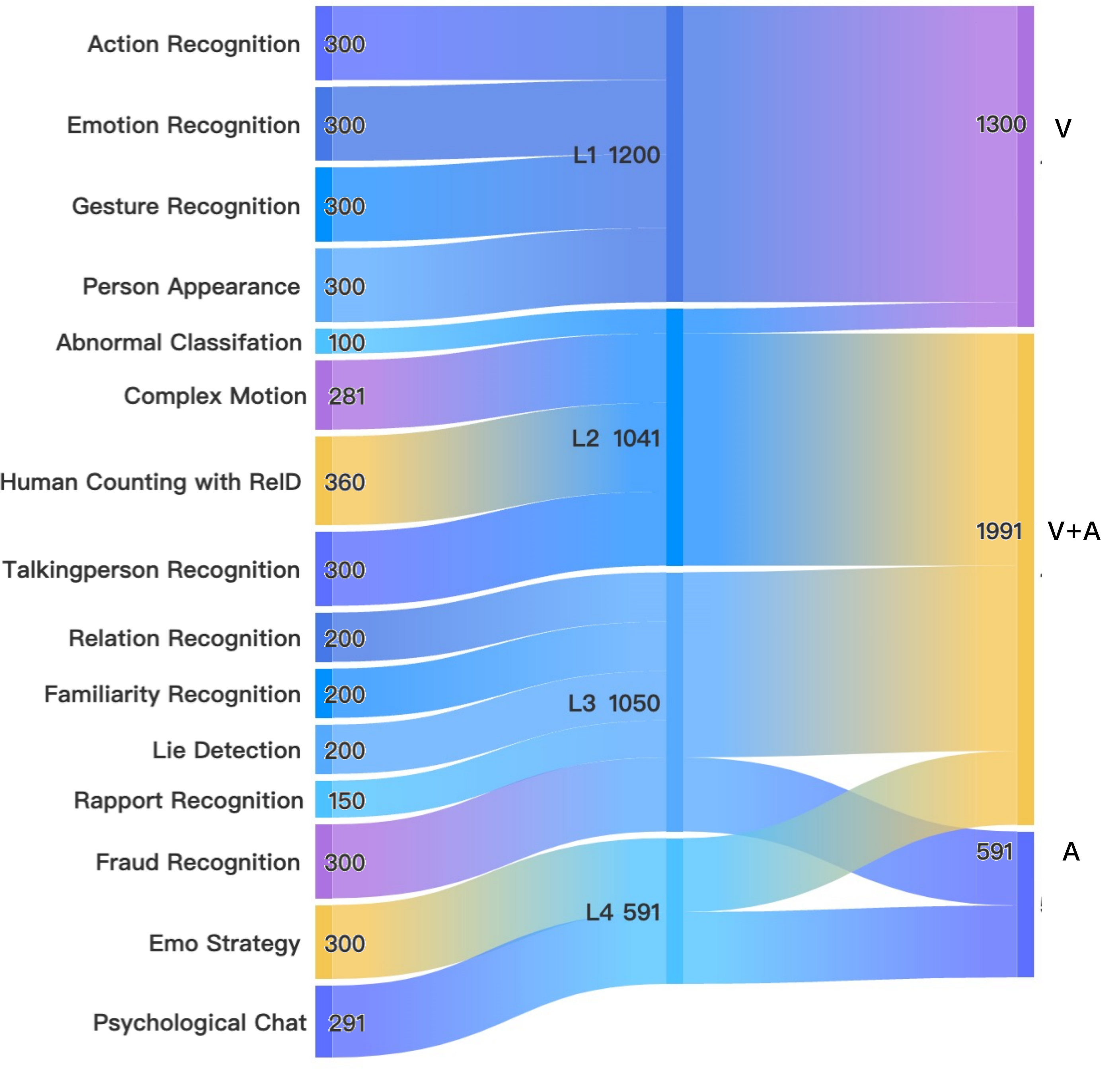} 
\caption{\textbf{Benchmark Statistics.}}
\label{task_count}
\end{figure}

\section{EVALUATION PROTOCALS}
During large model evaluation, besides video or audio inputs, we employ the following prompt to guide the model in answering multiple-choice questions:
\begin{figure}[H]
\centering
\includegraphics[width=1\columnwidth]{AnonymousSubmission/LaTeX/images/prompt_duoxuan.pdf}
\end{figure}
And we use the following prompt to have the model perform one-choice question:
\begin{figure}[H]
\centering
\includegraphics[width=1\columnwidth]{AnonymousSubmission/LaTeX/images/prompt_danxuan.pdf}
\end{figure}
And we use the following prompt to have the model perform open-ended question~(Human counting with Reid):
\begin{figure}[H]
\centering
\includegraphics[width=1\columnwidth]{AnonymousSubmission/LaTeX/images/prompt_wenda.pdf}
\end{figure}

For the trained model, we augment each question with a prompt designed to stimulate its reasoning process, which is as follows:
\begin{figure}[H]
\centering
\includegraphics[width=1\columnwidth]{AnonymousSubmission/LaTeX/images/prompt_think.pdf}
\end{figure}

\section{MORE DETAILS OF DATA CONSTRUCTION}
First, for a series of relatively straightforward tasks—including Person Appearance, Action Recognition, Gesture Recognition, Fraud Recognition, Emotion Recognition, Abnormal Classification, and Lie Detection—we construct multiple-choice questions based on their respective original datasets: CelebV-HQ, AVA, Jester,Telecom Fraud Texts, CelebV-HQ, ActivityNet, UCF-Crime100, and SEUMLD.

For each task, we first establish a comprehensive answer set. Each video clip's corresponding ground truth (GT) label serves as the correct option, while distractor options are randomly sampled from the remaining choices in the answer set. A multiple-choice question is then constructed by combining the video with the correct option and the sampled distractors.

\paragraph{Human Counting with ReID.}
We generate question-answer (QA) pairs for this task using the multi-object tracking dataset TAO. We first filter the TAO videos to exclude those without human subjects, then segment the remaining videos into 30-50 second clips.

For each segment, we identify the corresponding annotations from the original labels and calculate the total number of unique individuals present. This count serves as the numerical answer for our designed task, forming a complete question-answer pair.

\paragraph{Talking Person Recognition.}
We filter and construct questions based on videos from RealTalk.
First, we filter the clips, discarding any segments where only one person speaks continuously. For each selected video that contains a dialogue, we extract a single quote. This quote is then used to pose the question: ``Who said this?"
The objective of this task is to test a large model's multimodal comprehension ability, specifically its capacity to synchronize audio with visual information to identify the speaker.

\paragraph{Relation Recognition.}
The show "The Skin Deep" displays title cards at the beginning of each episode indicating the relationship between the two speakers. To create our question-answer pairs, we process the introductory segment of each video frame-by-frame using Optical Character Recognition (OCR) to extract this relationship information.

After filtering the raw OCR output to remove irrelevant text, the resulting clean text serves as the ground truth answer for the question: ``What is the relationship between the two people in the video?"

\paragraph{Familiarity Recognition.}
At the beginning of ``The Skin Deep" episodes, on-screen text often indicates the duration of the two speakers' acquaintance. We process these introductory segments frame-by-frame using Optical Character Recognition (OCR) to extract this information.

After filtering the raw OCR output, the resulting text serves as the ground truth answer for the question: "What is the level of familiarity between the two individuals in the video?"

\paragraph{Rapport Recognition.}
We segment clips from ``The Skin Deep" into 100-150 second lengths. We then conduct a two-level assessment of each segment using both the video and its corresponding Automatic Speech Recognition (ASR) transcript.

The first level of assessment evaluates the conversational interaction based on the following criteria:
Interactivity: The dialogue should feature frequent back-and-forth between participants. Points are deducted for long, awkward silences or extended pauses for thought.
Emotional Climate: Positive and shared emotions—such as happiness, humor, and relaxation—along with clear emotional resonance between the speakers are scored positively. Divergent or mismatched emotional states between participants result in a lower score.
Participation Balance: Contributions to the conversation should be roughly equal from both parties. Scenarios where one individual dominates the conversation while the other is primarily a passive listener are penalized.
Topical Coherence: The discussion should remain focused on a consistent topic. Frequent and abrupt switching between topics will negatively impact the score. The prompt is as follows:
\begin{figure}[H]
\centering
\includegraphics[width=1\columnwidth]{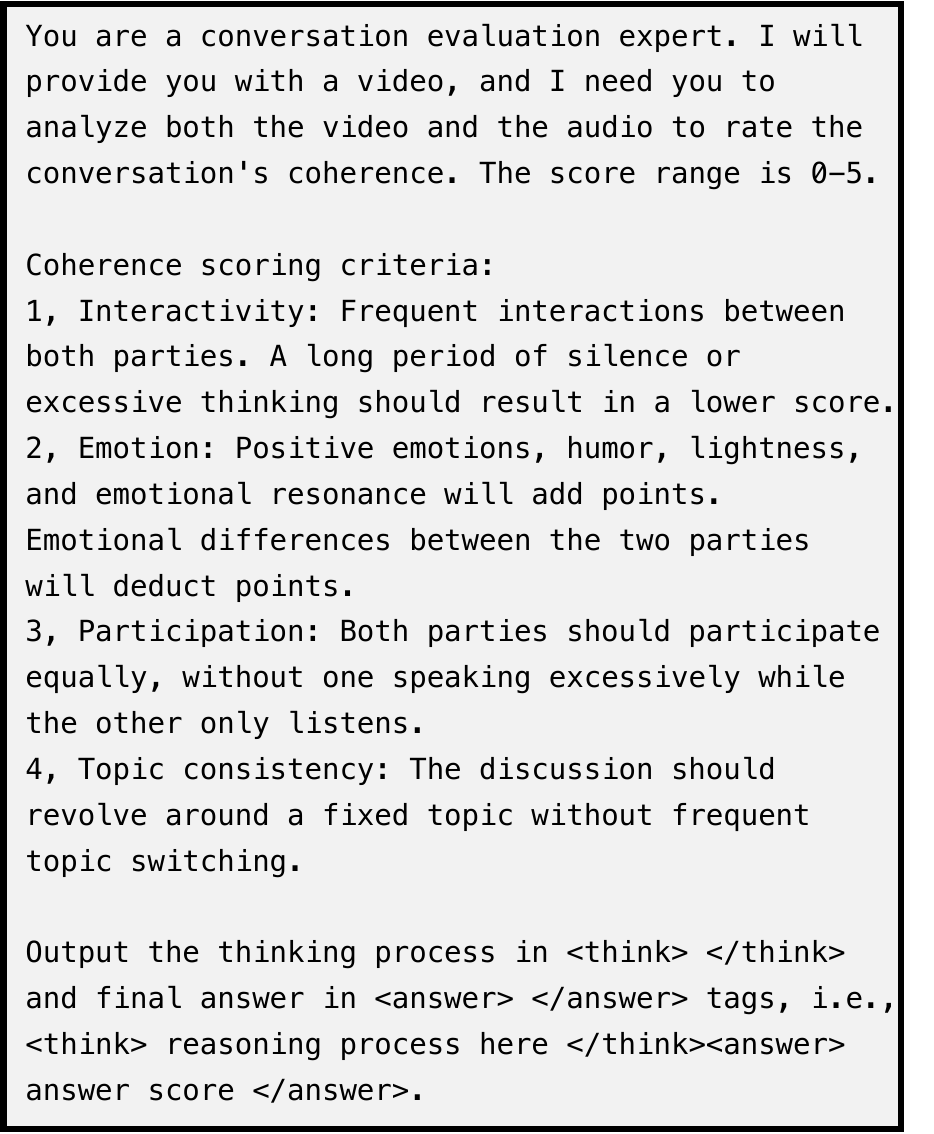}
\end{figure}

Secondly, we analyze the content of the conversation to determine if there are any disagreements.
The prompt is as follows:
\begin{figure}[H]
\centering
\includegraphics[width=1\columnwidth]{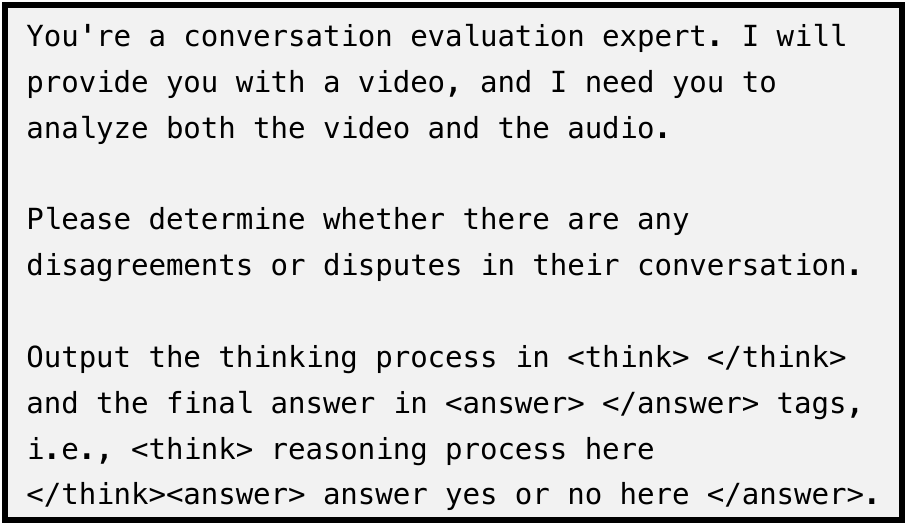}
\end{figure}


\paragraph{Emo Strategy.}
We use pre-segmented two-person conversation clips from the RealTalk dataset, which are originally used for training speaker generation models. For our task, we select the ``talker" from each clip as the input. The corresponding ground truth label is the ``listener"s' facial expression, which we determine using a facial expression recognition algorithm. We refer to this as our 'emotion strategy' target.

\paragraph{Psychological Chat.}
First, we filter for suitable dialogues to use as source material. This is done using the following prompt:
\begin{figure}[H]
\centering
\includegraphics[width=1\columnwidth]{AnonymousSubmission/LaTeX/images/prompt_chat1.pdf}
\end{figure}
Second, we retain the dialogues that score 5 or higher and then generate three distractor options for each one. This step is performed using the prompt below:
\begin{figure}[H]
\centering
\includegraphics[width=1\columnwidth]{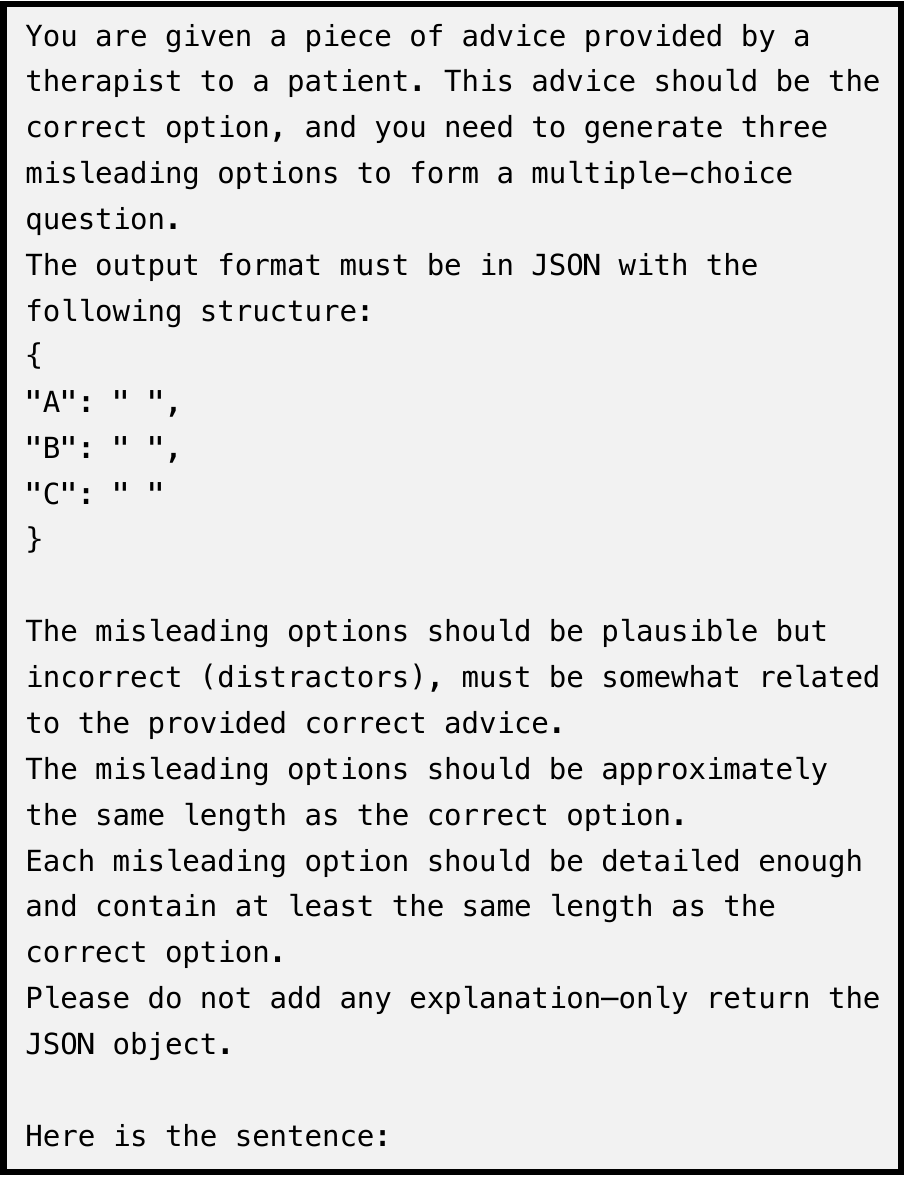}
\end{figure}

\begin{table*}[t]
\centering
\resizebox{1\textwidth}{!}{
\begin{tabular}{lllllllllllllllll}
\hline 
Models            & \textbf{Avg.} & AR    & ER    & GR    & PA    & AC    & CM    & HC    & TR    & FR1   & FR2   & LD    & RR1   & RR2   & ES    & PC    \\\hline 
Baseline          & 0.559         & 0.473 & 0.513 & 0.303 & 0.350 & 0.600 & 0.630 & 0.438 & 0.600 & 0.770 & 0.740 & 0.550 & 0.713 & 0.650 & 0.607 & 0.399 \\
First Stage (v)   & 0.584         & 0.693 & 0.493 & 0.427 & 0.608 & 0.530 & 0.626 & 0.438 & 0.597 & 0.720 & 0.557 & 0.540 & 0.707 & 0.620 & 0.593 & 0.540 \\
Second Stage (a)  & 0.601         & 0.700 & 0.510 & 0.407 & 0.600 & 0.550 & 0.630 & 0.492 & 0.587 & 0.775 & 0.687 & 0.545 & 0.693 & 0.625 & 0.593 & 0.625 \\
Third Stage (v+a) & 0.619         & 0.705 & 0.513 & 0.420 & 0.615 & 0.550 & 0.633 & 0.488 & 0.740 & 0.780 & 0.687 & 0.555 & 0.673 & 0.690 & 0.620 & 0.619\\\hline 
\end{tabular}
}
\caption{Evaluation of Optimization Results based on Qwen2.5-Omni-7B. We provide the scores for each task.}
\label{table3}
\end{table*}



\section{Supplement the section on MLLM benchmarks in related works}
With the development of multimodal large models, evaluation benchmarks~\cite{chen2024sharegpt4video,fu2025video,wang2024lvbench,li2024omnibench,zhang2024flash} have emerged. While they have been successful in general visual understanding, they rarely focus on the subtleties of human perception, comprehension, and planning. In the field of human behavior analysis, there are some specialized datasets. For example, in emotion recognition, AffectNet~\cite{mollahosseini2017affectnet} and in action recognition, Kinetics~\cite{smaira2020short}. While these datasets are crucial, they are typically designed for specific classification or detection tasks and do not provide a framework for evaluating a model's comprehensive reasoning ability. HumanOmniV2~\cite{yang2025humanomniv2} focuses on deciphering intentions, interpreting emotions, and detecting potential deception from video.
In summary, existing benchmarks either lack multimodal information, focus on general scenarios rather than human-centric ones, or are too task-specific. To systematically evaluate the capabilities of large models in complex human-centered scenarios, we have developed HumanSense.

\bigskip

\bibliography{aaai2026}